\documentclass[10pt,journal,compsoc]{IEEEtran}

\ifCLASSOPTIONcompsoc
  \usepackage[nocompress]{cite}
\else
  \usepackage{cite}
\fi

\usepackage{subcaption}
\usepackage{enumitem}
\usepackage{epsfig}
\usepackage{graphicx}
\usepackage{multirow} 

\usepackage[pagebackref=true,breaklinks=true,letterpaper=true,colorlinks,bookmarks=false]{hyperref}

\begin{document}

\title{Diverse Semantic Image Editing with \\ Style Codes}

\author{
Hakan Sivuk and Aysegul Dundar

\IEEEcompsocitemizethanks{
\IEEEcompsocthanksitem  H. Sivuk and A. Dundar are with Department of Computer Science, Bilkent University, Ankara, Turkey
}
}


\IEEEtitleabstractindextext{
\begin{abstract}
Semantic image editing requires inpainting pixels following a semantic map.
It is a challenging task since this inpainting requires both harmony with the context and strict compliance with the semantic maps.
The majority of the previous methods proposed for this task try to encode the whole information from erased images.
However, when an object is added to a scene such as a car, its style cannot be encoded from the context alone. 
On the other hand, the models that can output diverse generations struggle to output images that have seamless boundaries between the generated and unerased parts.
Additionally, previous methods do not have a mechanism to encode the styles of visible and partially visible objects differently for better performance. 
In this work, we propose a  framework that can encode visible and partially visible objects with a novel mechanism to achieve consistency in the style encoding and final generations. 
We extensively compare with previous conditional image generation and semantic image editing algorithms. 
Our extensive experiments show that our method significantly improves over the state-of-the-art. 
Our method not only achieves better quantitative results but also provides diverse results. Please refer to the project web page for the released code and demo:  \href{https://github.com/hakansivuk/DivSem}{https://github.com/hakansivuk/DivSem}.

\end{abstract}
}

\maketitle
\IEEEdisplaynontitleabstractindextext
\IEEEpeerreviewmaketitle

\section{Introduction}

Semantic image editing refers to the task of generating consistent pixels with the input image based on semantic conditioning. 
The generated parts need to be in harmony with the rest of the image while strictly following the content provided as semantic maps.
This task enables adding objects, manipulating scenes, and erasing objects as well as outpainting and extending images. 
It is also referred to as conditional inpainting and is considered to be highly challenging as the inpainted region should be coherent with the overall image and the provided semantic map.

While semantic image editing previously required manual effort as in many other image editing and inpainting tasks, today promising results are achieved with deep learning-based generative methods without any human intervention \cite{ntavelis2020sesame, luo2022context}.
One biggest challenge of semantic image editing is to keep the context style consistent between inpainted and unerased regions and achieve seamless integration of those regions.
There have been previous methods proposed for semantic-based conditional image generation. 
Those methods, such as SPADE \cite{park2019semantic}, can generate an image from scratch but do not have a mechanism to encode styles for visible and generated pixels. The same semantic-based style encoding is fed to instance normalization layers, and when their architecture is trained for semantic image editing task, this results in unsatisfactory image editing results.  
Another work, SEAN \cite{zhu2020sean}, learns styles for each semantic region while generating images. However, SEAN does not have a mechanism to encode partially erased images.
Therefore, it cannot be used efficiently for an inpainting task.

There are other works proposed for semantic image editing task, such as
SESAME \cite{ntavelis2020sesame} and SPMPGAN \cite{luo2022context}.
SESAME also uses a SPADE architecture and does not provide a mechanism to encode valid and invalid pixels differently or a mechanism for style encoding. 
SPMPGAN demonstrates improved semantic image editing results over SESAME. They propose a style-preserved modulation layer that includes a two-stage modulation method, the first to inject the style information from the semantic map and the second to inject the encoded styles coming from the feature maps.
Their work also does not include a mechanism for style encoding.
ASSET \cite{liu2022asset} proposes a transformer-based approach that can achieve high-resolution image editing results. 
However, it suffers from boundary artifacts due to the inconsistencies across generated and existing parts, especially in crowded scenes such as Cityscapes and ADE20k-Room images \cite{luo2023siedob}. 
SIEDOB \cite{luo2023siedob} shows that diverse object addition can be achieved but it is through separately trained multiple sub-networks and only for a few selected foreground objects and not for semantic classes.
In our work, to achieve correct style encoding for partially erased and completely erased semantic classes, we propose a novel end-to-end framework and show that our method achieves better results than those previous works.

\begin{figure*}[]
\includegraphics[width=\textwidth]{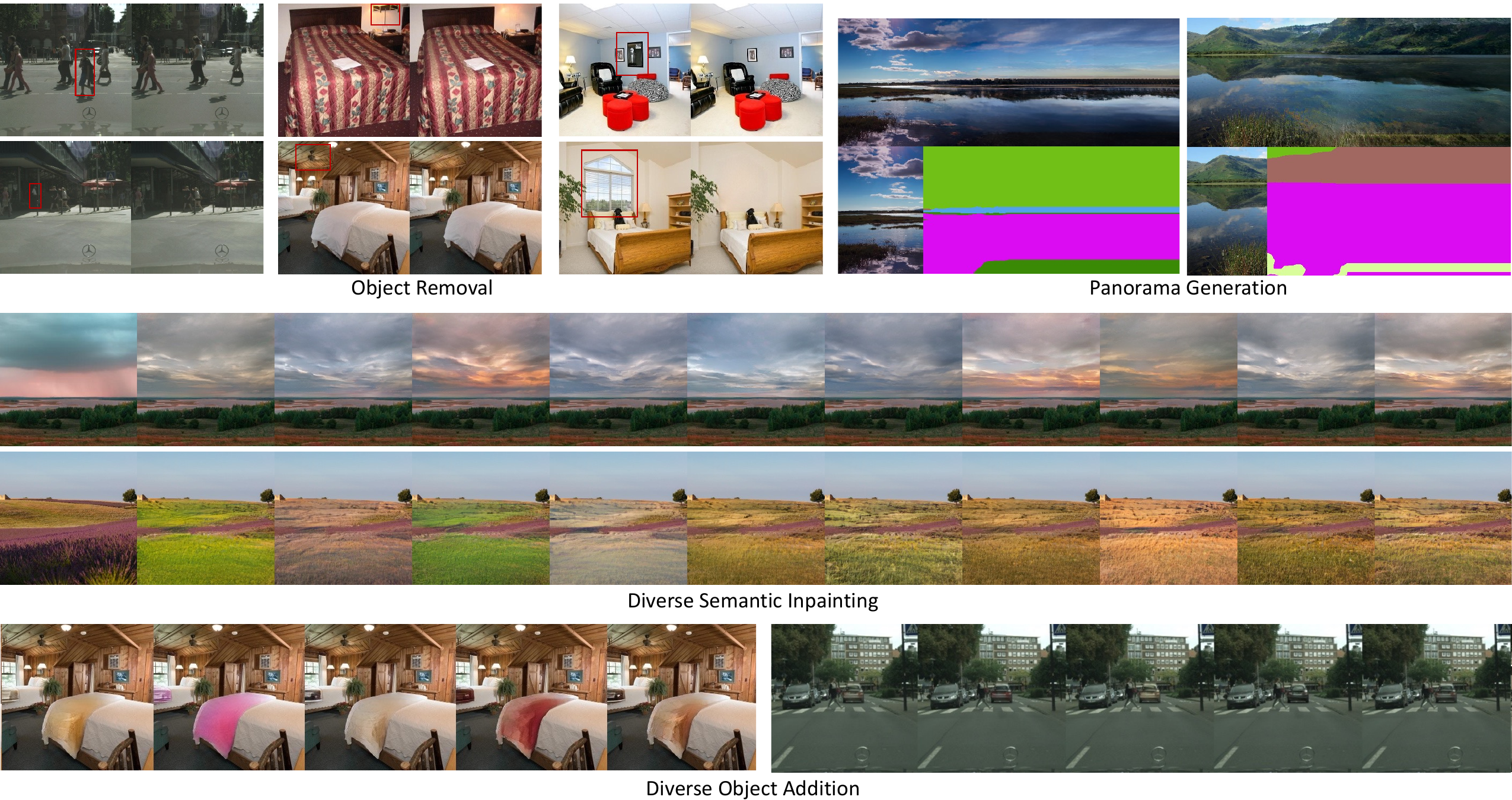}
\caption{Our method can remove objects from a scene (highlighted with red boxes), generate panoramas and inpaint a semantic class with different styles, and add objects to a scene again in different styles.}
\label{fig:teaser}
\end{figure*}

Furthermore, even though previous works have achieved significant progress, none of them can generate both diverse semantic image editing outputs and consistency across generated and unerased parts. 
Most of these models try to encode all the information to generate an image from the erased input images and semantic maps.
However, when a new object, e.g. a car, is added to a scene or a semantic class, e.g. sky, is erased completely to be regenerated, the erased input image does not carry the style information for successful generations. 
These models struggle with these scenarios and do not have the ability to copy styles from reference images.
Achieving diversity via style encodings from reference images is also another advantage of our work.
Our framework, as shown in Fig. \ref{fig:teaser}, can remove objects from a scene, add objects in diverse styles, generate panoramas, and inpaint a semantic class again with different styles. 
It also achieves inpaintings of free-form masks that have arbitrary shapes.
In summary, our main contributions are:
\begin{itemize}[leftmargin=*]
    \item We propose a novel training pipeline that combines the encoded styles from erased images for visible parts and styles from complete images if a semantic category or instance is completely erased. This way, the reconstruction objective can be satisfied during training, and different styles can be extracted from different images to achieve diverse results during the inference.
    \item We propose a mask-aware module for encoding the styles of partially erased semantic classes or instances to achieve style extraction consistency across different sizes of masks.
    \item We compare our method with several state-of-the-art conditional image inpainting methods. We provide results with an extensive number of metrics. Our results show the effectiveness of our framework with significant improvements over state-of-the-art. 
    \item We show applications of our framework that include object removal, diverse object addition, panorama generations, and diverse semantic inpainting. 
\end{itemize}

\section{Related Work}

Image-to-image translation algorithms find many applications such as image colorization \cite{dong2020cycle}, image upsampling \cite{kim2020unsupervised}, texture prediction \cite{dundar2022fine, dundar2023progressive, altindis2023refining}, and editing image domains \cite{starganv2, xiao2018elegant, dalva2022vecgan,park2019semantic, dundar2020panoptic, zhu2020sean, ntavelis2020sesame, liu2022partial, dalva2023image}.
In image-to-image translation methods, generations can be conditioned on different inputs such as depth \cite{zhang2023adding, esser2021taming}, keypoints \cite{dundar2020unsupervised, wang2021one}, edge maps \cite{wang2018high}, and semantic maps \cite{park2019semantic, dundar2020panoptic, zhu2020sean, ntavelis2020sesame, liu2022partial}  to name a few. 
Among them, conditioning image generation on to semantic maps is one of the most popular because those maps are easy to paint, and they provide pixel-level semantic information and therefore provide good user control to the generation process.
This task was dominated by Generative Adversarial Networks (GANs) \cite{park2019semantic, dundar2020panoptic, zhu2020sean, ntavelis2020sesame, liu2022partial}  until recently.
In the last couple of years, diffusion models empowered with large training sets of text-image pairs also show significant results on semantically conditioned image generation \cite{rombach2022high, wang2022pretraining}.
These models additionally take text prompts \cite{singh2022high, avrahami2022spatext,wang2022pretraining,rombach2022high} to describe the appearance of objects which may not be optimal to explain the coloring and style information of parts.
Therefore, it is arguable that some of the proposed GAN-based methods \cite{zhu2020sean} provide better user control by relying on the appearance encoding from reference images.

In this work, we are interested in semantic image editing task in which image editing is conditioned on segmentation maps.
This task requires successful image inpainting with generated pixels following the semantic map conditioning \cite{ntavelis2020sesame, luo2022context}.
Therefore, our work is related to image inpainting methods as well.
Extensive research is conducted on image inpainting task that includes inpainting with GAN inversion \cite{yildirim2023diverse, yu2022high, wang2022dual, tov2021designing, pehlivan2023styleres}.
These works require StyleGAN models \cite{karras2019style} trained on millions of images for different objects such as faces and cars. 
It may not be always scalable. 
In our work, we show successful results with training on datasets as small as two thousand images.
There have been other works that propose partial \cite{liu2018image, liu2022partial} and gated \cite{yu2019free} convolutions and mask-aware transformers \cite{li2022mat} to handle the erased pixels and valid ones differently. 
Our style encoding module has the same motivations as these works.
We additionally take advantage of the semantic or instance maps to encode styles for visible and invisible parts differently. 
Our work has more capabilities than those inpainting works since we can edit an image by modifying the semantic shapes of the objects, changing the style of the instances, removing objects from the scene, and outpainting of images with semantic control. 

Semantic image editing has an additional challenge than conditional image generation or unconditional image inpainting since the generated pixels need to be consistent with unerased/unmodified input pixels and semantic maps. 
Successful applications of this task are shown with GAN-based methods \cite{ntavelis2020sesame, luo2022context}.
However, these models are not multi-modal and try to generate novel objects in the scene by encoding visible parts.
Therefore, they do not have a mechanism to generate diverse images. 
Recently, ASSET \cite{liu2022asset} proposes a transformer-based architecture that is built on VQGAN \cite{esser2021taming}.
Via sparse attention mechanisms and using a codebook in an autoregressive setting, they can output diverse images. 
However, ASSET suffers from boundary artifacts between the visible and inpainted regions \cite{luo2023siedob, liu2022asset}. 
SIEDOB  \cite{luo2023siedob} proposes generating foreground and background pixels separately and then fusing them at the end. 
It requires training an instance inpainting network, a style-diversity object generator, and a background generator separately and in the end, a fusion network to combine the outputs of previous networks.
SIEDOB can generate diverse results for foreground objects but not for semantic categories and requires many sub-networks and separate training for each.
Our method is trained end-to-end and can generate diverse foreground and background classes.

Lastly, diffusion-based image inpainting methods have been extensively studied recently \cite{lugmayr2022repaint, rombach2022high, yildirim2023inst}, but they do not provide a semantic control on the generation.
To the best of our knowledge, there is only one work of semantic editing with a diffusion-based method \cite{goel2023pair}. 
This concurrent work shows that the structure of content can be edited with diffusion models thanks to the separately encoded semantic maps and appearance codes. 
However, rather than inpainting, their model requires generating a semantic class from scratch, and even when the partial content is erased, there is no guarantee that the valid pixels stay the same. The valid pixels can be carried to the output image with the help of a mask, however, this will cause inconsistencies between generated and valid/unmasked pixels. Therefore, it is not comparable with our work. 
In this work, we propose a novel architecture and training pipeline to achieve diverse semantic editing. The training objective includes image reconstruction and adversarial losses.
The architecture has a novel way of encoding styles from reference and input images.

\begin{figure*}[h!]
\includegraphics[width=\textwidth]{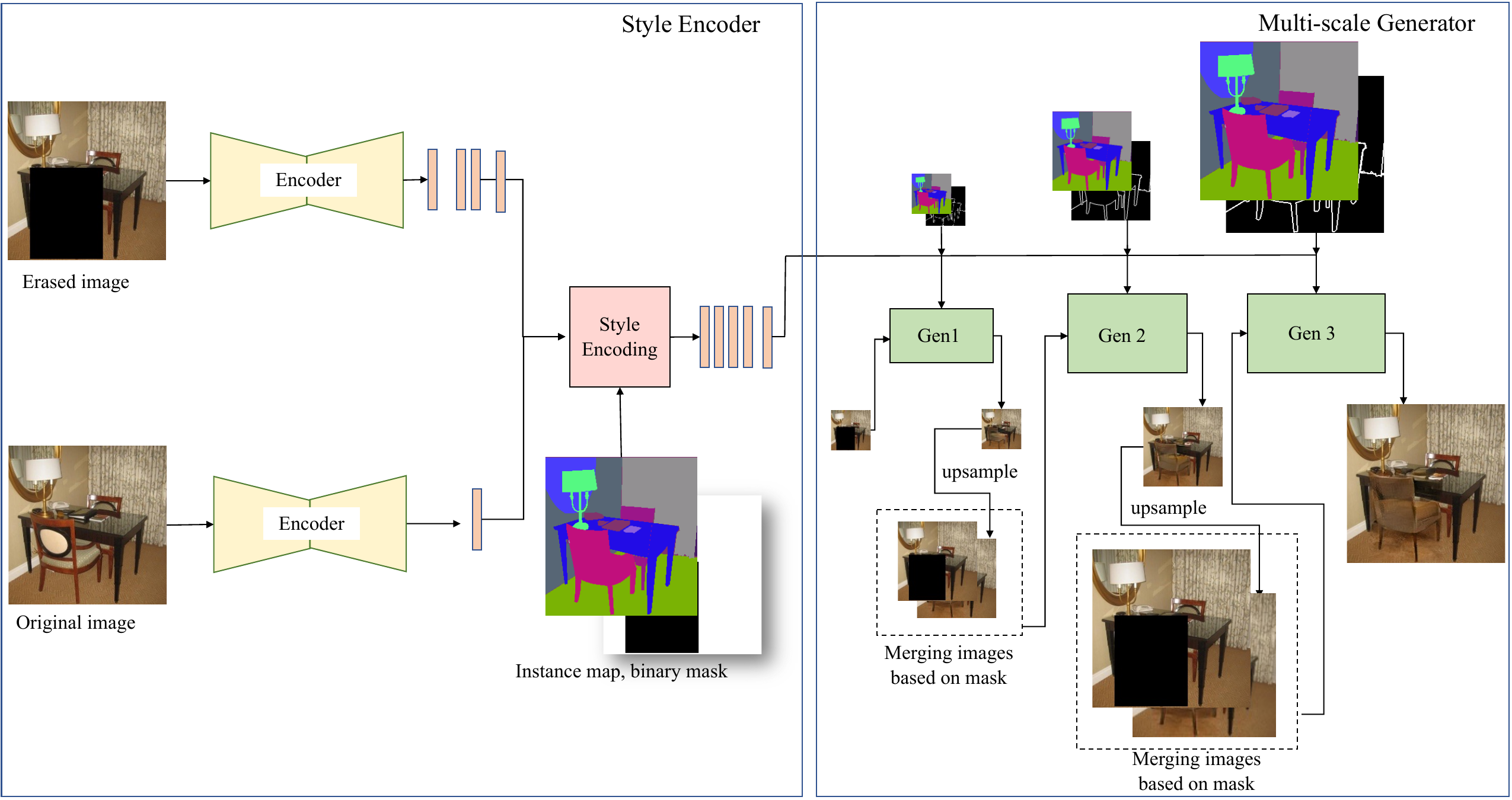}
\caption{The overall pipeline for the diverse semantic image editing. 
During training, we both encode the erased and original images. The original image is encoded because if a semantic or instance ID is completely erased, then its style is extracted from the original image for the training.
The encoded styles, instance maps, and binary mask are fed into the style encoding module which is explained in more detail in Sec. \ref{sec:style_encoding}. 
The outputs of the style encoding are used in the normalization layers of the generator.
The normalization layers additionally take semantic maps, edge maps, and binary masks as inputs.
As for the generator, we use a multi-scale generator that refines the predictions at each stage as explained in more depth in Sec. \ref{sec:generator}.}
\label{fig:diagram}
\end{figure*}

\section{Method}

Our method includes a novel style encoding pipeline, training that facilitates style learning, and a generator that outputs the target images. 
We explain each of these modules in this section.

\subsection{Style Encoding}
\label{sec:style_encoding}

The goal of the style encoding module is to encode styles from erased images.
However, encoding an erased image alone may not be always enough. 
For example, if a semantic class for example the whole sky, or an instance such as a car is erased completely, then there is no information about the styles of these parts in the erased images.
Previous works such as SPMPGAN \cite{luo2022context} and SESAME \cite{ntavelis2020sesame} try to encode everything from the erased images. 
In our work, we encode those styles from the original image during training as shown in Fig. \ref{fig:diagram}.

Our style encoder is inspired by SEAN \cite{zhu2020sean} but includes significant differences to adapt the style encoding idea for the inpainting task.
SEAN encodes styles from unerased images for the image generation task. 
It employs an encoder that includes symmetric convolutional downsampling and upsampling layers.
The output of the encoder is spatially the same size as the input image.
Based on the semantic maps for stuff (e.g. sky, road) and instance maps for objects (e.g. person, car), a region-wise averaging pooling is applied which was also used in pix2pixHD \cite{wang2018high}.
This pooling results in a bottleneck by removing the spatial information from the encoded features.
The encoded styles broadcasted to the original resolution based on the semantic and instance maps which are used in the adaptive normalization layers in the generator.

In our setting, the pixels that correspond to a semantic or instance ID can be completely erased,  partially erased, or not erased at all. 
For the not-erased and partially-erased parts, we encode the styles from erased input images. 
However, when the input images are masked, some pixels are 0 at the beginning. Therefore, using these features that correspond to the masked pixels in the average operation of instance-wise pooling affects the calculations. So, if we directly use instance-wise pooling in the image inpainting domain, it will cause inconsistency in the style code calculation.

To make style code calculation consistent across different masking ratios, we propose an algorithm where the average operation is only performed on the valid area corresponding to the object, which means we only take the mean of the values from the valid area.
Additionally, we learn a normalization layer that is based on the valid ratio of the region. 
For example, if 100 pixels out of 500 pixels are erased for the sky pixels, the valid ratio is 0.2.
We use this valid ratio as an input to the two $1\times1$ linear layers followed by sigmoid operation which gives the scale and shift that will be applied to the style codes. 
With this mechanism, the network can handle the partially-erased and not-erased parts differently as they carry different information.
The shift and scale can be set to ineffective values (shift=0, scale=1) with sigmoid operation by the network if the valid ratio information is found not useful.
However, we observe that the network chooses to use this information and results improve. We show these results in our Ablation studies.

There might be also some objects that are fully covered with the mask. In that case, the number of valid pixels is 0 for these objects and semantic maps. 
For this scenario,  during the training, we use the same input image but without any mask, as can be shown in Fig. \ref{fig:diagram}. This is to encode the style of that object or semantic class. We use the rest of the styles from the erased input image. 
In other words, we encode style code from the original image and perform instance-wise pooling.
The valid ratio is 1 for these cases and the normalization is applied again.
With this algorithm, calculated style codes are numerically more consistent with the masked input image and fully masked objects do not degrade training performance since no information would be encoded to the style code for these objects.
Note that in these tasks, previous works \cite{ntavelis2020sesame, luo2022context} set the objective for the training as full image reconstruction even though the input does not carry the information to reconstruct the target completely.
Our setting provides access to the full information of the image to the network and can work better with the complete image reconstruction objective. 

For inference time, we provide styles for the fully erased objects and semantic maps from our training images.
This also allows us to generate images diversely with reference styles.

To sum up, in the Style encoding module shown in Fig. \ref{fig:diagram}, the module takes instance and binary maps and styles encoded from erased and original images. If there exists a semantic or instance ID with zero valid pixel, for that object a style is taken from the original image.
The Style encoding module also does the normalization based on the valid pixel ratio. 
The binary mask and instance maps are given to the style encoding module to calculate the valid pixel ratio and to detect if an instance or semantic class is completely erased.
The encoded styles are used in the generator which is explained in the next section.

Finally, the architectural details of the encoder that is used in the style encoder are as follows: 
For the encoder, following a $3\times3$ convolution, we use 4 successive blocks that perform downsampling, which reduces feature map dimensions to $16\times16$. After these downsampling blocks, we perform a $3\times3$ convolution to increase channel size to 512. In our decoder, we perform 4 upsampling by a factor of 2 where each is followed by a $3\times3$ convolution. Skip connections coming from the encoder are concatenated with feature maps between upsampling and convolution layers. Finally, a $3\times3$ convolution is performed to set channel size to style length. Except for this final convolution block, we use instance normalization denoted as (+IN). The channels increase as \{16, 32, 64, 128, 256, 512\} in the encoder and decrease as \{512, 256, 256, 256, style-length\}.
The output of this encoder goes through the instance-wise pooling operation.
The style-length is set to $128$ and other alternatives are reported in Ablation Studies.

\subsection{Generator}
\label{sec:generator}

We use a multi-scale generator that is based on SPMPGAN \cite{luo2022context}. It consists of a pyramid of generators; Gen1, Gen2, and Gen3. 
The input resolutions that go to the style encoding and Gen3 are $256\times256$. 
That includes the erased images ($I_3$), binary ($M_3$), and semantic maps ($S_3$). 
Additionally, our target image is in $256\times256$ resolution.
However, in Gen1, we start by generating a smaller image. 
Specifically, the erased input image, mask, and semantic maps are downsampled to $64\times64$ which we refer to as $I_1, M_1, S_1$.
The output of Gen1, $O_1$, is upsampled and combined with $I_2$ based on the visibility masks, $O_1 \cdot M_1 + I_2 \cdot (1-M_1)$.  
This part is shown as merging images based on the mask in Fig. \ref{fig:diagram}.

This generated image goes to Gen2 and after a similar procedure, Gen3 outputs our final image.
All of these generators have an encoder-decoder structure with skip connections to remove the information bottleneck. 
By incorporating skip connections, the decoder can access higher-resolution features from the encoder, which can help preserve fine details and contextual information.  

Erased and intermediate-generated images are fed as input to the convolutional architecture in the generators. 
The encoded styles, semantic maps, and edge maps are fed via normalization layers. 
We encode the styles only once that are used across multi-scale generators.
The normalization layer we use integrates context encoding \cite{luo2022context}, semantic layout, binary mask, and encoded styles. Styles are broadcasted to the original input resolution based on the semantic and instance IDs of pixels.
Semantic layouts, binary masks, edge maps, and styles are concatenated at first via the channel dimension.
Two groups of parameters are encoded from them by convolutional layers, $(\gamma_{s1}, \beta_{s1})$ and $(\gamma_{s2}, \beta_{s2})$.
The context modulation parameters, $(\gamma_{c}, \beta_{c})$, are obtained from feature maps before the instance normalization is applied.  
This way the context styles that are encoded through the convolutional layers are carried by the feature maps via the context modulation \cite{luo2022context}.
These two modulations are fused as follows:

\begin{equation}
    \gamma_f = (1+\gamma_{s2}) \cdot \gamma_{c} + \beta_{s2} 
\end{equation}
\begin{equation}
    \beta_f = (1+\gamma_{s1}) \cdot \beta_{c} + \beta_{s1} 
\end{equation}

These shift $\beta_f$ and scale  $\gamma_f$ parameters are used after the instance normalization is applied to the features. 

The architectural details of the generators are as follows: 
For the encoder, following a $3\times3$ convolution, we use 5, 6, and 7 successive convolution blocks respectively for Gen1, Gen2, and Gen3 that perform downsampling. Each downsampling reduces feature map dimensions by 2. After these downsampling blocks, we perform a final $3\times3$ convolution in the encoder. In the decoders, we have 5, 6, and 7 successive convolution blocks respectively for Gen1, Gen2, and Gen3. Finally, a $3\times3$ convolution is performed to set channel size to 3 which corresponds to the RGB images. Inside these convolution blocks, after increasing feature map dimensions by 2, skip connections coming from the encoder are concatenated with feature maps. Concatenated feature maps are given to the gated convolution layer. Except for this final convolution block, we use contextual and style normalization in addition to instance normalization denoted as (+IN). For Gen 1, the channels increase as \{32, 64, 128, 128, 256, 256, 512\} in the encoder and decrease as \{256, 128, 64, 32, 32, 3\} in the decoder. For Gen 2, the channels increase as \{32, 64, 128, 128, 256, 256, 512, 512\} in the encoder and decrease as \{512, 256, 128, 64, 32, 32, 3\} in the decoder. For Gen 3, the channels increase as \{32, 64, 128, 128, 256, 256, 512, 512, 1024\} in the encoder and decrease as \{512, 256, 256, 128, 64, 32, 32, 3\} in the decoder.

\newcommand{\interpfiga}[1]{\includegraphics[width=2.4cm,height=2.4cm]{#1}}
\begin{figure*}[t]
  \centering
  \setlength\tabcolsep{1pt}
  \renewcommand{\arraystretch}{1}
  \begin{tabular}{ccccccc}
  \multirow{2}{*}{\rotatebox{90}{ADE20k-Room}}  &
    \interpfiga{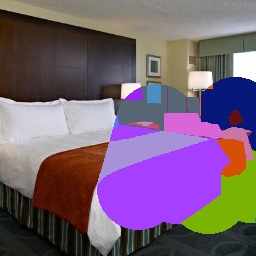}&
    \interpfiga{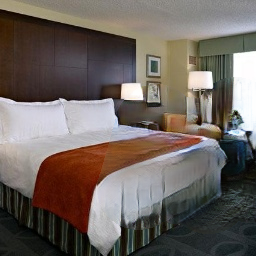}&
    \interpfiga{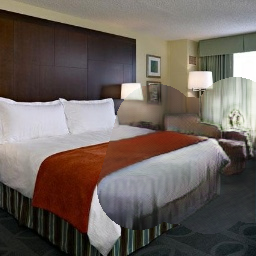}&
    \interpfiga{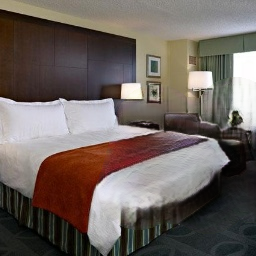}&
    \interpfiga{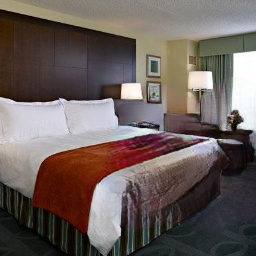}&
    \interpfiga{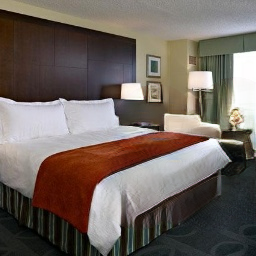} \\
   &
    \interpfiga{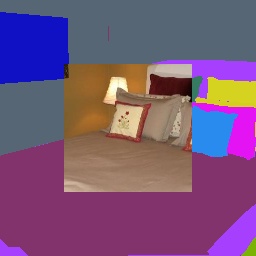}&
    \interpfiga{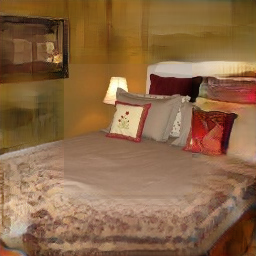}&
    \interpfiga{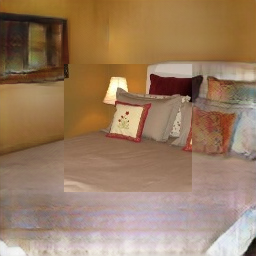}&
    \interpfiga{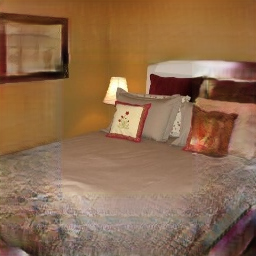}&
    \interpfiga{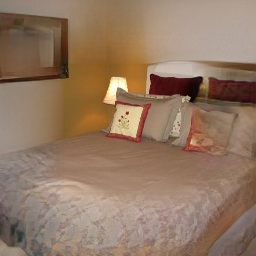}&
    \interpfiga{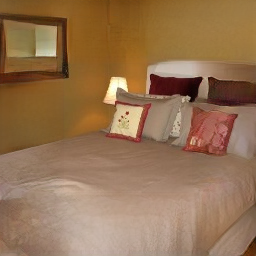} \\
    \multirow{2}{*}{\rotatebox{90}{Ade20k-Landcape}} &
    \interpfiga{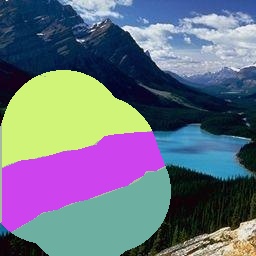}&
    \interpfiga{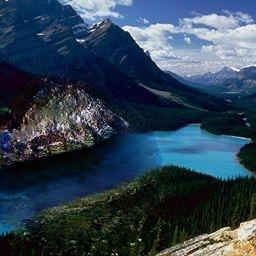}&
    \interpfiga{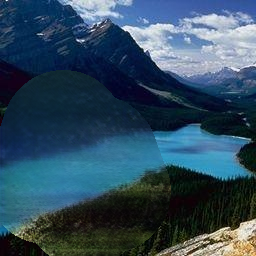}&
    \interpfiga{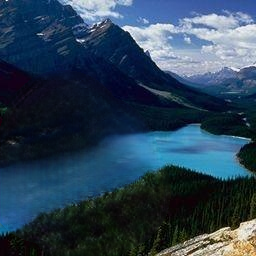}&
    \interpfiga{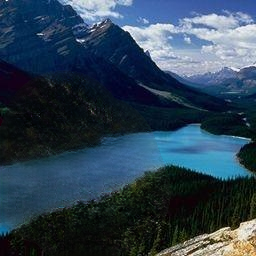}&
    \interpfiga{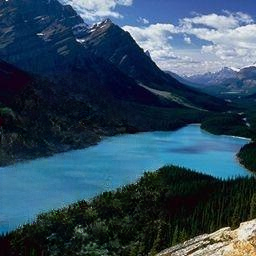} \\

    &
    \interpfiga{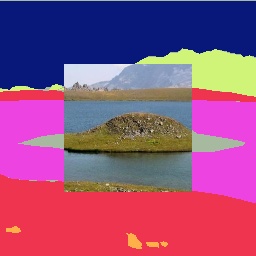}&
    \interpfiga{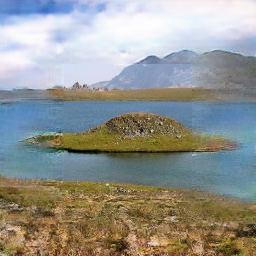}&
    \interpfiga{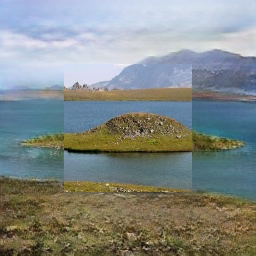}&
    \interpfiga{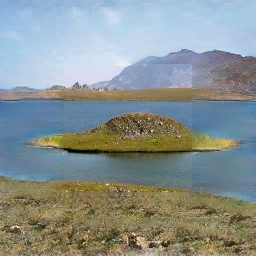}&
    \interpfiga{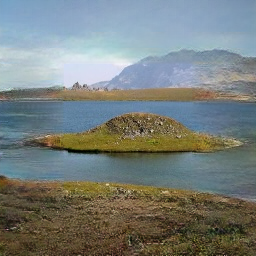}&
    \interpfiga{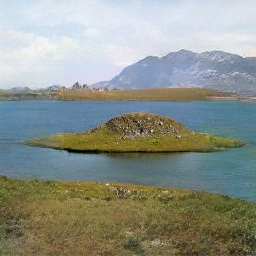} \\
    & Input+Mask & SPADE & SEAN & SESAME & SPMPGAN & Ours

     \end{tabular}
     \caption{Comparisons of methods on ADE20k dataset for free-form and outpainting masks. Our method achieves better results in terms of color consistency between the generated and existing pixels and sharpness in the generated parts.}
     \label{fig:comparisons_masks}
\end{figure*}

\begin{figure*}[t]
  \centering
  \setlength\tabcolsep{1pt}
  \renewcommand{\arraystretch}{1}
  \begin{tabular}{cccccccc}
    
     \multirow{3}{*}{\rotatebox{90}{Cityscapes}}  &  \interpfiga{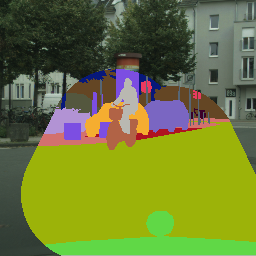}&
     \interpfiga{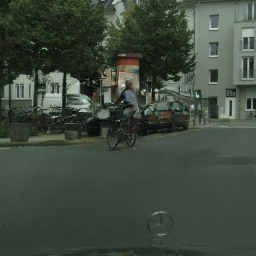}&
    \interpfiga{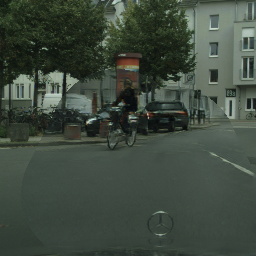}&
    \interpfiga{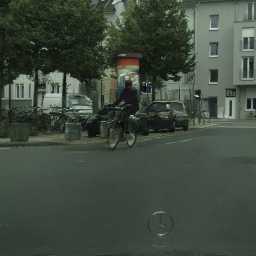}&
    \interpfiga{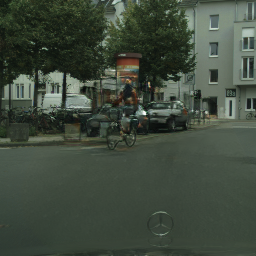}&
    \interpfiga{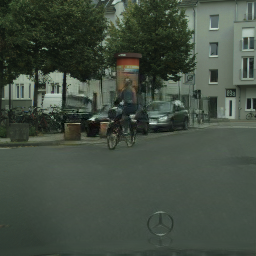}&
    \interpfiga{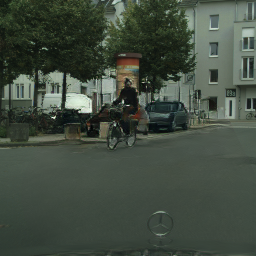} \\
    &  
    \interpfiga{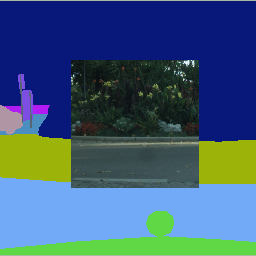} & 
    \interpfiga{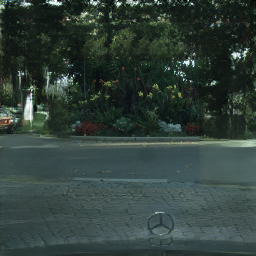} &
    \interpfiga{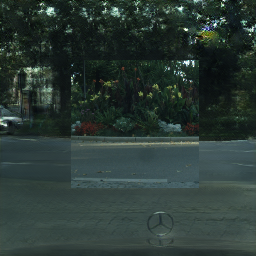} &
    \interpfiga{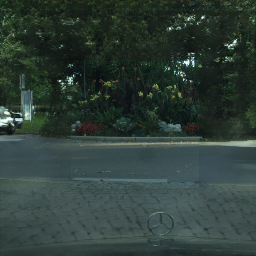} &
    \interpfiga{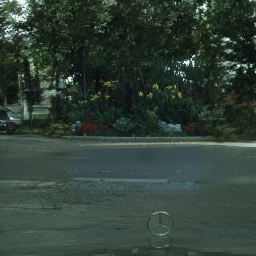} &
    \interpfiga{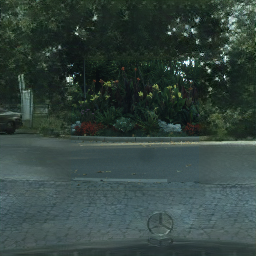} &
    \interpfiga{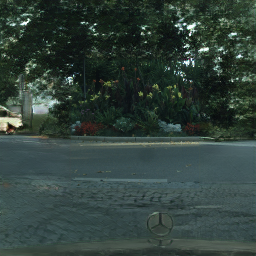} \\
    &  
    \interpfiga{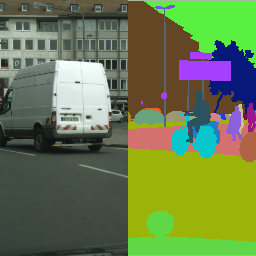} & 
    \interpfiga{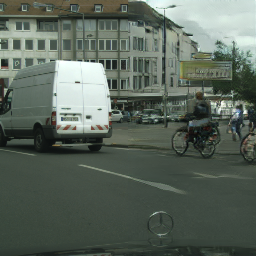} &
    \interpfiga{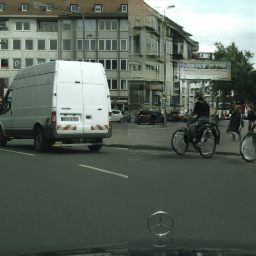} &
    \interpfiga{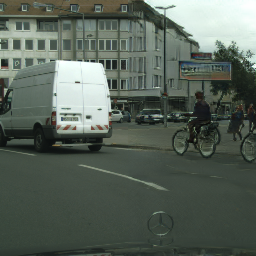} &
    \interpfiga{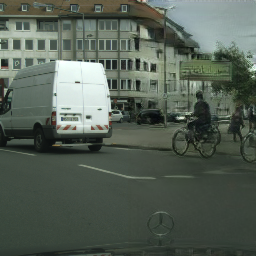} &
    \interpfiga{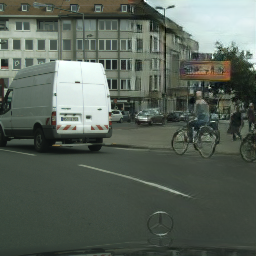} &
    \interpfiga{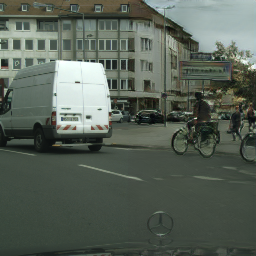} \\

    & Input+Mask & SPADE & SEAN & SESAME & SPMPGAN & SIEDOB & Ours
 
     \end{tabular}
     \caption{Comparisons of methods on Cityscapes datasets for free-form, outpainting, and extension masks. Our method achieves better results in terms of color consistency between the generated and existing pixels and sharpness in the generated parts.}
     \label{fig:comparisons_masks_city}
\end{figure*}

\subsection{Training Parameters}  

Our model is trained end-to-end using a combination of adversarial and reconstruction loss functions. 
For adversarial learning, we attach a discriminator for each of the generators. 
The discriminators include convolutional layers with a filter size of $5\times5$. 
The smallest discriminator that guides Gen1 has 4 layers, and the others have 5 and 6 layers, respectively. 
Inside the discriminator, channel size increases from 64 by a factor of 2 at each layer until it reaches 512. In the end, it decreases to 3. 
Leaky ReLU is used as an activation function and stride is used as 2.

As for the loss functions, we employ adversarial (\(L_{adv}\)), perceptual loss (\(L_P\)), and L1 distance losses (\(L_1\)) at each scale.  The overall loss function used for training the model is given by:
\[L = L_{adv} + 10 \times L_P + L_1\]

We use these parameters following SPMPGAN \cite{luo2022context} and do not tune them. We achieve better scores by changing the architectures without tuning any parameters.
We train our model on a single GPU for 500 epochs with batch size 3, \(\beta_{1}=0.5\), \(\beta_{2}=0.999\), weight decay 0.0001, learning rate 0.0001.

\begin{table*}[t]
\centering
\caption{Comparisons of our framework with the state-of-the-art methods. We highlight the best in \textbf{bold} and the second best with \underline{underline}. ADE20k-Landcape does not have instances, so we do not calculate the scores for that task on the ADE20k-Landcape dataset.}
\begin{tabular}{|l|l|l|l||l|l|l||l|l|l||l|l|l|}
\hline 
&  \multicolumn{12}{|c|}{Cityscapes}   \\
\hline
& \multicolumn{3}{c||}{ Free-Form }& \multicolumn{3}{c||}{ Extension }& \multicolumn{3}{c||}{ Outpainting}  & \multicolumn{3}{c|}{Adding Objects}  \\
\hline
Methods  & FID & mIoU & Acc. &  FID & mIoU & Acc. &  FID & mIoU & Acc. & FID & mIoU & Acc. \\
\hline
SPADE \cite{park2019semantic} & 18.70  & \underline{57.94} & 93.77 & 33.10 & 56.47 & 93.63 & 41.37 & 55.99 & 93.65 & 3.92 & 58.14 & 93.80 \\
SEAN \cite{zhu2020sean} & 21.27  & 57.81  & \underline{93.79} & 35.38 & 56.92 & \underline{93.87} & 43.06 & 56.08 & 93.70 & 4.08 & 58.42 & 93.81 \\
SESAME \cite{ntavelis2020sesame} & \underline{17.60}  & \textbf{59.16} & \textbf{94.07} & \underline{32.06} & \textbf{59.66} & \textbf{94.25} & \underline{39.94} & \textbf{60.86} & \textbf{94.36} & \underline{3.67} & 58.55 & \textbf{93.84} \\
SPMPGAN \cite{luo2022context}  & 18.76 & 57.14 & 93.61 & 34.53 & 55.93 & 93.55 & 41.88 & 56.52 & 93.55 & 3.74 & \textbf{58.91} & \textbf{93.84} \\
SIEDOB \cite{luo2023siedob} & 18.71 & 56.78 & 93.60 & 34.33 & 56.14 & 93.53 & 43.21 & \underline{56.54} & 93.46 & 4.17 & 58.65 & 93.79 \\
\hline
Ours & \textbf{17.03} & 57.89 & 93.72 & \textbf{31.05} & \underline{57.09} & \underline{93.73} &  \textbf{39.53} & 56.44 & \underline{93.73} & \textbf{3.49} & \underline{58.87} & \underline{93.83}\\
\hline
&  \multicolumn{12}{|c|}{ADE20k-Room}   \\
 \hline

SPADE  \cite{park2019semantic}  & 22.64 & 25.70 & 82.39 & 37.63 & 25.91 & 82.07 & 47.69 & 26.00 & 81.72 & 17.00 & 26.81 & 81.65\\
SEAN  \cite{zhu2020sean} & 25.74 & 26.52 & 81.84 &  36.92 & 25.21 & 81.09 &  47.05 & 25.07 & 81.11 & 15.39 & 27.63 & 82.03 \\
SESAME \cite{ntavelis2020sesame} & 21.35 & \underline{27.94} & \textbf{83.43} & 34.77 & \underline{27.94} & \underline{83.52} & 44.61 & \underline{28.54} & 83.84 & 15.33 & 28.48 & 82.56 \\
SPMPGAN \cite{luo2022context} & \underline{19.09} & 27.29 & 82.96 & \underline{32.63} & 27.70 & 83.36 & \underline{40.82} & 28.28 & \underline{84.02} & \underline{14.46} & \underline{28.68} & \textbf{82.88} \\
\hline
Ours & \textbf{18.40} & \textbf{28.09} & \underline{83.38} & \textbf{32.05} & \textbf{28.07} & \textbf{83.80} & \textbf{40.13} & \textbf{28.71} & \textbf{84.47} & \textbf{13.50} & \textbf{28.80} & \underline{82.80} \\
\hline
&  \multicolumn{12}{|c|}{ADE20k-Landscape}   \\
 \hline
SPADE  \cite{park2019semantic}  & 32.51 & 28.89 & 75.02 & 59.62 & 24.51 & 74.80 & 78.61 & 22.82 & 75.81 & - & - & - \\
SEAN  \cite{zhu2020sean} & 44.96 & 27.35 & 74.00 &  66.68 & \underline{24.12} & 73.91 &  89.97 & 21.66 & 74.78 & - & - & - \\
SESAME \cite{ntavelis2020sesame} & 31.88 & 28.95 & 75.29 & 57.15 & 26.38 & 75.00 & 74.38 & \underline{26.81} & \textbf{78.59} & - & - & - \\
SPMPGAN  \cite{luo2022context} & \underline{25.22} & \underline{29.10} & \textbf{75.78} & \underline{52.03} & 26.26 & \textbf{75.82} & \underline{69.29} & 25.15 & 77.45 & - & - & - \\
\hline
Ours & \textbf{24.19} & \textbf{29.24} & \underline{75.52} & \textbf{49.28} & \textbf{24.08} & \underline{75.19} & \textbf{64.80} & \textbf{27.16} & \underline{78.43} & - & - & - \\
\hline
\end{tabular}
\label{table:results_all}
\end{table*}

\section{Experiments}

\subsection{Datasets}

We run experiments on three datasets.

\textbf{ADE20k-Room.} Following \cite{luo2022context},  we select a subset of the ADE20k dataset \cite{zhou2017scene} comprised of Bedroom, Hotel Room, and Living Room. This subset is called ADE20k-Room. This dataset has 2246 images for training and 225 for testing.
ADE20k dataset provides 150 semantic labels, however, on this subset, we end up with 117 semantic labels.

\textbf{ADE20k-Landscape.} We select the landscape subclass from ADE20k dataset for our experiments following  \cite{luo2022context}. The training set and the testing set contain 1695 images and 153 images, respectively.
This data subset has 81 unique semantic classes.

\textbf{Cityscapes.} We conduct our experiments on Cityscapes \cite{cordts2016cityscapes} datasets that have both instance and semantic segmentation labels available. The Cityscapes dataset contains 2,975 training images and 500 validation images of urban street scenes along with 35 semantic classes and 9 instance classes.

\subsection{Evaluation Masks}

To assess the model, we create masks that are pre-determined for testing and used for the evaluation of all models. Various types of masks exist, each serving a distinct purpose in evaluating the model from different angles.

\textbf{Free-Form.}
Unlike regular masks, which are usually predefined shapes such as rectangles or circles, free-form masks can have any arbitrary shape, making them more flexible. This allows for realistic and visually coherent inpainting results, as the model learns to generate content that is consistent with the surrounding context of the image. 

\textbf{Extension.} Extension masks are utilized to eliminate the left portion of an image, requiring the model to produce the missing pixels using only the information available in the right half of the image. 
This setting enables the generation of panoramas as well. 

\textbf{Outpainting.} These masks are designed to keep the central quarter of an image, forcing the model to generate the extensions in the borders using only the center parts of the image.

\textbf{Adding Objects.} These masks are produced through the process of randomly selecting an instance within the image, with the objective of covering this selected instance using the smallest possible rectangular mask.
Adding objects is used for Cityscapes and ADE20k-Room datasets since they include various instances. On the other hand, this setting does not apply to the ADE20k-Landscape dataset which consists of semantic classes. 

\begin{figure*}[t]
  \centering
  \setlength\tabcolsep{1pt}
  \renewcommand{\arraystretch}{1}
  \begin{tabular}{ccccccc}
  \multirow{2}{*}{\rotatebox{90}{ADE20k-Room}}  &
    \interpfiga{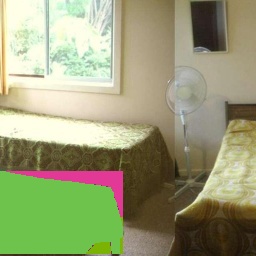}&
    \interpfiga{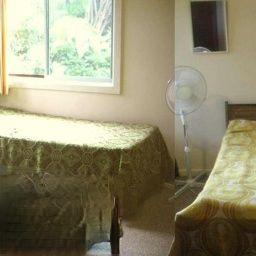}&
    \interpfiga{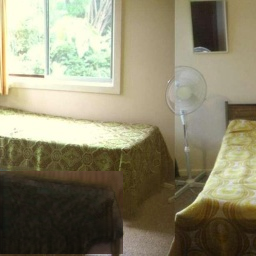}&
    \interpfiga{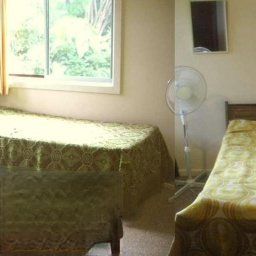}&
    \interpfiga{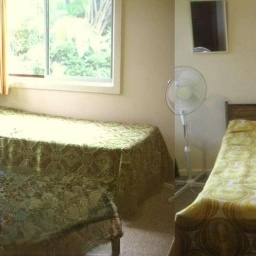}&
    \interpfiga{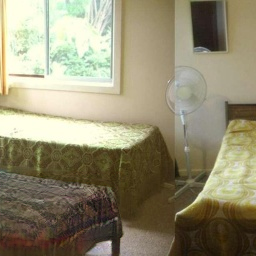} \\
   &
    \interpfiga{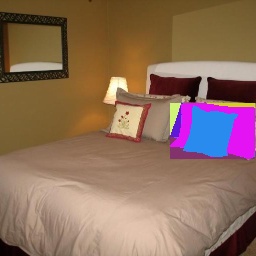}&
    \interpfiga{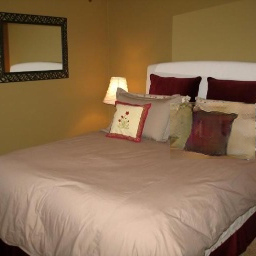}&
    \interpfiga{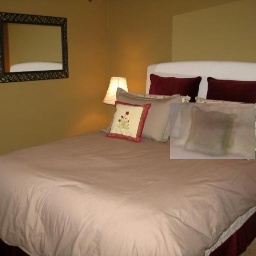}&
    \interpfiga{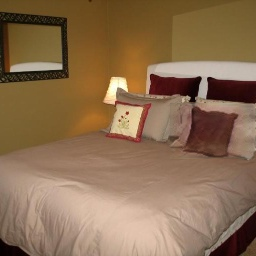}&
    \interpfiga{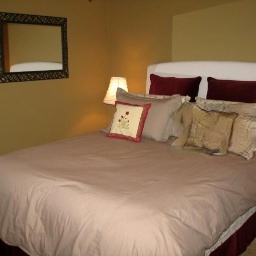}&
    \interpfiga{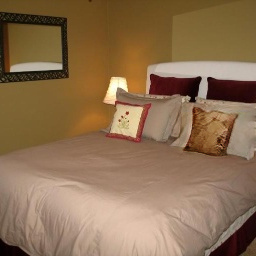} \\
    &
    \interpfiga{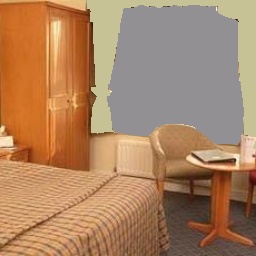}&
    \interpfiga{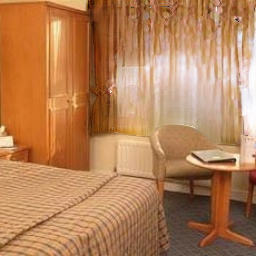}&
    \interpfiga{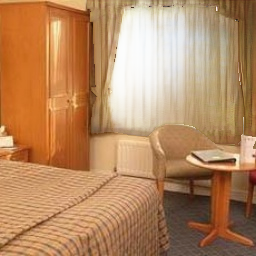}&
    \interpfiga{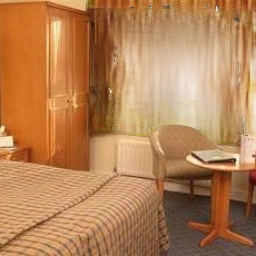}&
    \interpfiga{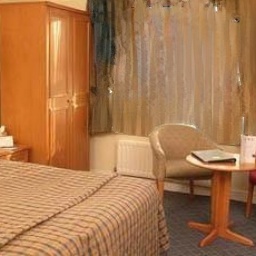}&
    \interpfiga{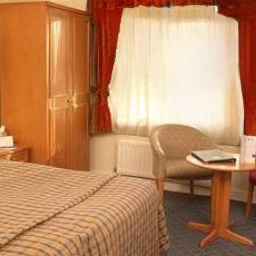} \\
    & Input+Mask & SPADE & SEAN & SESAME & SPMPGAN & Ours

     \end{tabular}
     \caption{Comparisons of methods on  ADE20k-Room dataset for adding objects setting. Our method can add objects with diverse styles which improves realism.}
     \label{fig:comparisons_add_ade}
\end{figure*}

\begin{figure*}[t]
  \centering
  \setlength\tabcolsep{1pt}
  \renewcommand{\arraystretch}{1}
  \begin{tabular}{cccccccc}
    
     \multirow{4}{*}{\rotatebox{90}{Cityscapes}}  &  \interpfiga{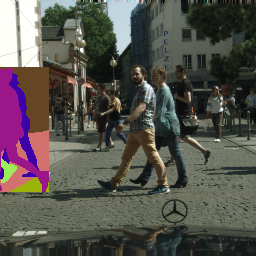}&
     \interpfiga{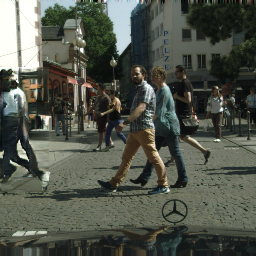}&
    \interpfiga{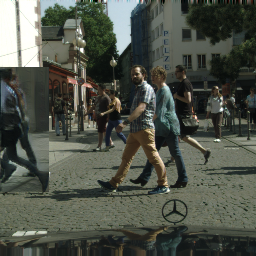}&
    \interpfiga{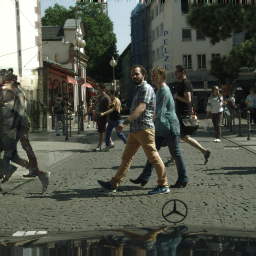}&
    \interpfiga{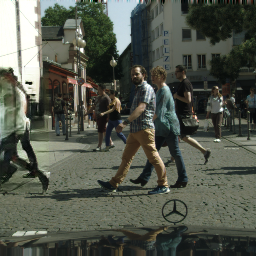}&
    \interpfiga{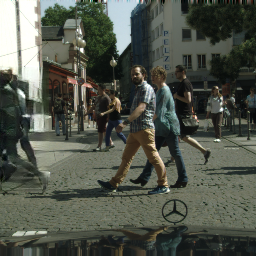}&
    \interpfiga{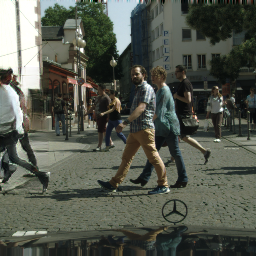} \\
    &  
    \interpfiga{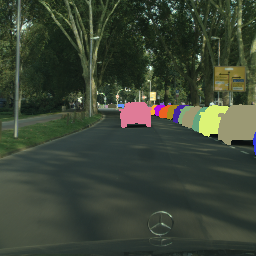} & 
    \interpfiga{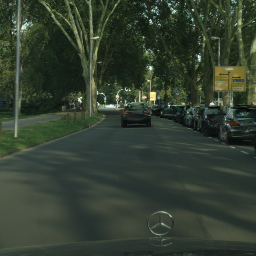} &
    \interpfiga{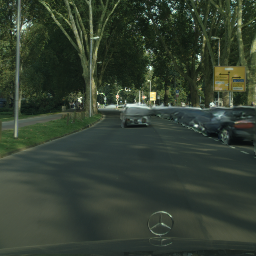} &
    \interpfiga{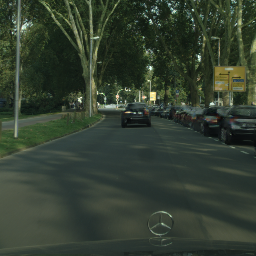} &
    \interpfiga{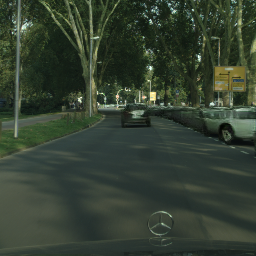} &
    \interpfiga{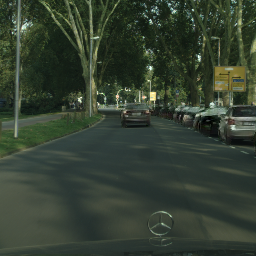} &
    \interpfiga{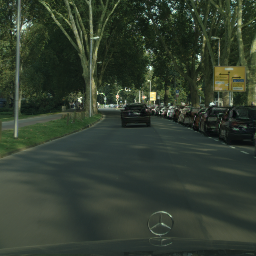} \\
    &  
    \interpfiga{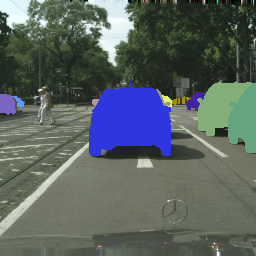} & 
    \interpfiga{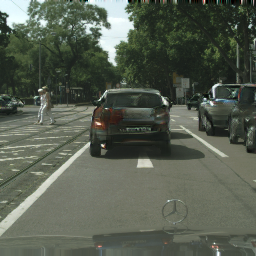} &
    \interpfiga{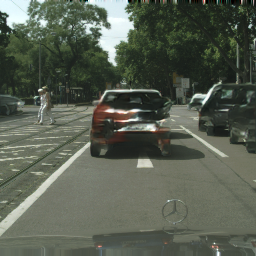} &
    \interpfiga{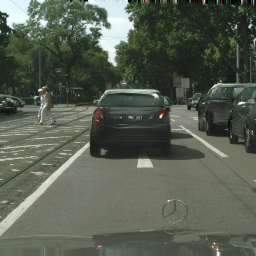} &
    \interpfiga{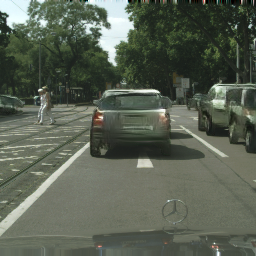} &
    \interpfiga{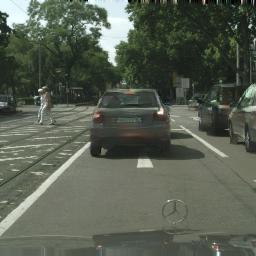} &
    \interpfiga{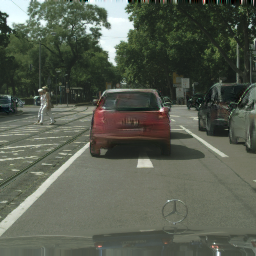} \\
    &  
    \interpfiga{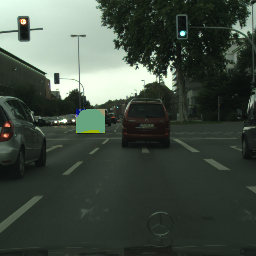} & 
    \interpfiga{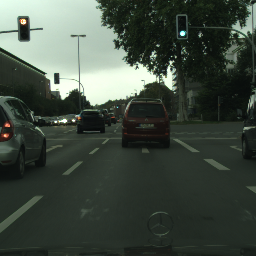} &
    \interpfiga{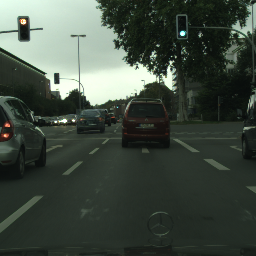} &
    \interpfiga{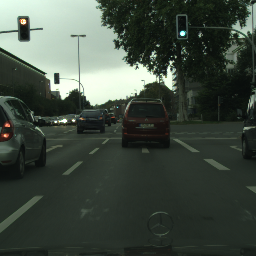} &
    \interpfiga{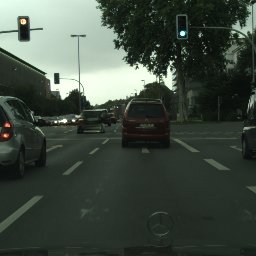} &
    \interpfiga{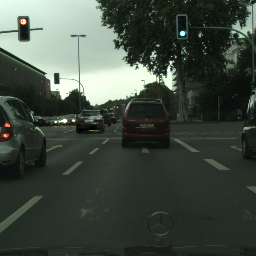} &
    \interpfiga{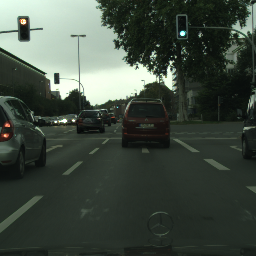} \\
    & Input+Mask & SPADE & SEAN & SESAME & SPMPGAN & SIEDOB & Ours
 
     \end{tabular}
     \caption{Comparisons of methods on Cityscapes dataset for adding objects setting. Our method can add objects with diverse styles which improves realism.}
     \label{fig:comparisons_add_city}
\end{figure*}

\subsection{Baselines}

We compare our method with SPADE \cite{park2019semantic}, SEAN \cite{zhu2020sean}, SESAME \cite{ntavelis2020sesame}, SPMPGAN \cite{luo2022context}, SIEDOB \cite{luo2023siedob}.
SPADE and SEAN are proposed for image generation tasks. 
We train them for inpainting by inputting the erased images and setting the target as the unerased image.

SESAME, SPMPGAN, and SIEDOB are proposed for the semantic-based image inpainting task. 
They take erased images as input. 
SESAME uses a similar architecture to SPADE whereas SPMPGAN uses a multi-scale architecture with a context-preserving normalization layer. 
They are both deterministic and cannot output diverse images.
SIEDOB proposes to handle foreground and background pixels differently. 
It requires training an instance inpainting network, a style-diversity object generator, and a background generator separately and in the end, a fusion network to combine the outputs of previous networks.
SIEDOB can generate diverse results for foreground objects but not for semantic categories.
We evaluate provided pretrained models of SPMPGAN and SIEDOB on our validation images.
For the other methods, we train them on our datasets. We use the released code of the authors.

\subsection{Metrics}
The performance of the proposed image inpainting model is evaluated using three widely adopted metrics for this task \cite{ntavelis2020sesame} and a user study. The first one is the Fréchet Inception Distance (FID) and the others are the mean Intersection over Union (mIoU) and Accuracy (Acc.). 

\textbf{FID} quantifies the distance between the distribution of the generated and real images in terms of their feature representations extracted by a pretrained Inception network \cite{heusel2017gans}. The smaller distance is considered to show a better quality in the generation.

\textbf{mIoU} and \textbf{Acc.} scores are computed by running a well-trained segmentation model on the generated images. 
For the Cityscapes dataset, we use DRN model \cite{yu2017dilated} and for the ADE20k datasets, we use UperNet \cite{xiao2018unified}.
The predicted maps from the segmentation networks are compared with the ground-truth labels to measure mIoU and pixel-level accuracy. 
With this metric, we measure how much the generator can follow the conditional segmentation map. 

\textbf{User Study} is conducted on 30 samples among 10 users. To generate 30 samples, the first 10 images of the validation set from each dataset are used without cherry-picking.  We set an A/B test and provide users with input images and generated ones obtained by our method and SPMPGAN.
We limit our user study to this comparison given the expenses of user study and since SPMPGAN outperforms the previous works.
The left-right order is randomized to ensure fair comparisons.
Users are asked if the generated parts are in harmony with the rest of the image while strictly following the content provided as semantic maps.

\subsection{Results}

We provide the quantitative results in Table \ref{table:results_all} for Cityscapes, ADE20k-Room, and ADE20k-Landcape datasets.
SIODEB releases pretraiend models on the Cityscapes dataset and is included in our quantitative results.
All methods achieve reasonable and similar mIoU and accuracy results.
This shows that methods can follow the semantic content. 
On the other hand, they generate different quality outputs as measured in FIDs.
Methods that are proposed for semantic-based image generation, SPADE and SEAN, achieve significantly worse results than the others in terms of FIDs. 
SESAME, SPMPGAN, and SIEDOB achieve better scores. 
Ours achieves the best FID scores consistently among all datasets and all mask forms. 
These results are in line with the visual results.

Qualitative results for free-form and outpainting masks on ADE20k-Room and ADE20k-Landcape images are shown in Fig. \ref{fig:comparisons_masks}.
SPADE, SEAN, and SESAME output images with obvious boundaries between the visible and inpainted pixels. 
SPMPGAN achieves better blending between the generated pixels and existing ones.
However, there are still inconsistencies in many examples.
As can be seen from the figure, our method achieves the best style preservation of partially erased semantic maps.
For example, in the third example, our method fills the rest of the lake with a similar color to the existing pixels from the lake while the others cannot. 
We additionally show free-form and outpainting mask results on Cityscapes images in Fig. \ref{fig:comparisons_masks_city}.
Our method achieves better color consistency and more distinct foreground objects such as the cyclist in the first row.

\begin{table}[t]
\centering
\caption{Ablation Study. FID results are given for different configurations. We highlight the best in \textbf{bold}.}
\begin{tabular}{|l|l|l|l|l|}
\hline 
\multicolumn{2}{|l|}{} & \multicolumn{3}{|c|}{ADE20k-Landcape}   \\
\hline
\multicolumn{2}{|l|}{Methods}  & Free-Form & Extension & Outpainting \\
\hline
\multicolumn{2}{|l|}{w/o Style} & 25.22 & 52.03 &  69.29   \\
\hline
\multicolumn{2}{|l|}{w/o Valid Ratio} & 25.95 & 51.23 &  66.58   \\
\hline
\multicolumn{2}{|l|}{w/o Multi-scale Gen.} & 27.03 & 52.48 & 69.38 \\
\hline
Style Dim. & 64 & 25.22  & 50.12 & 66.20   \\
Style Dim.  & 256 & 25.14  & 50.62 & 66.27   \\
\hline
\multicolumn{2}{|l|}{Ours}  & \textbf{24.19}  &  \textbf{49.28} &\textbf{64.80}   \\
\hline
\end{tabular}
\label{table:ablation}
\end{table}
 
\begin{figure*}[]
\includegraphics[width=\textwidth]{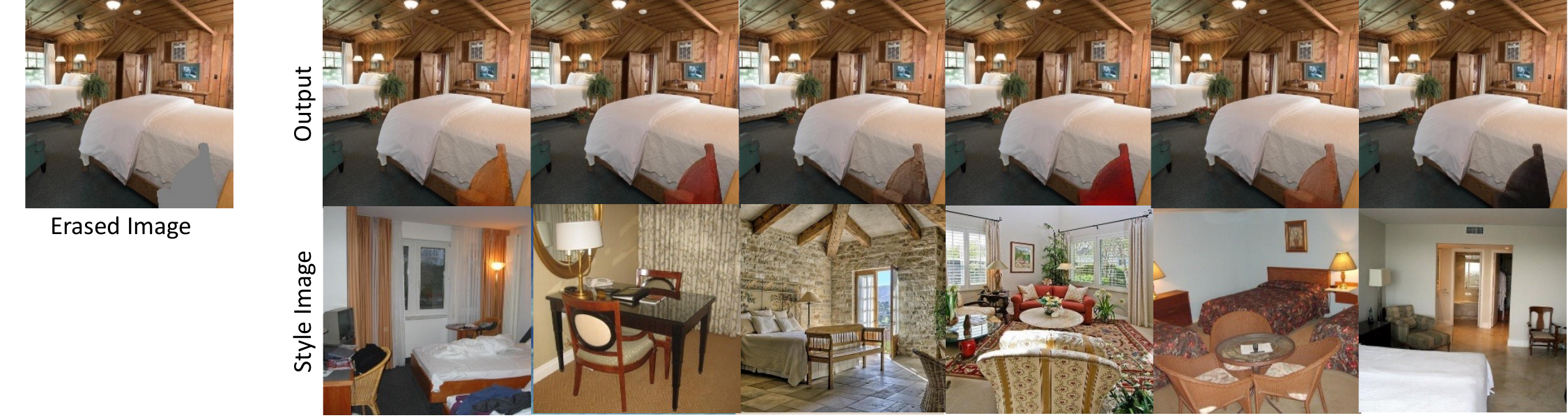}
\caption{Adding diverse objects with our method. The chair is erased from the image and is inpainted with different style codes that are extracted from corresponding semantic classes of the style images.}
\label{fig:diverse_objects}
\end{figure*}

\begin{figure*}[]
\includegraphics[width=\textwidth]{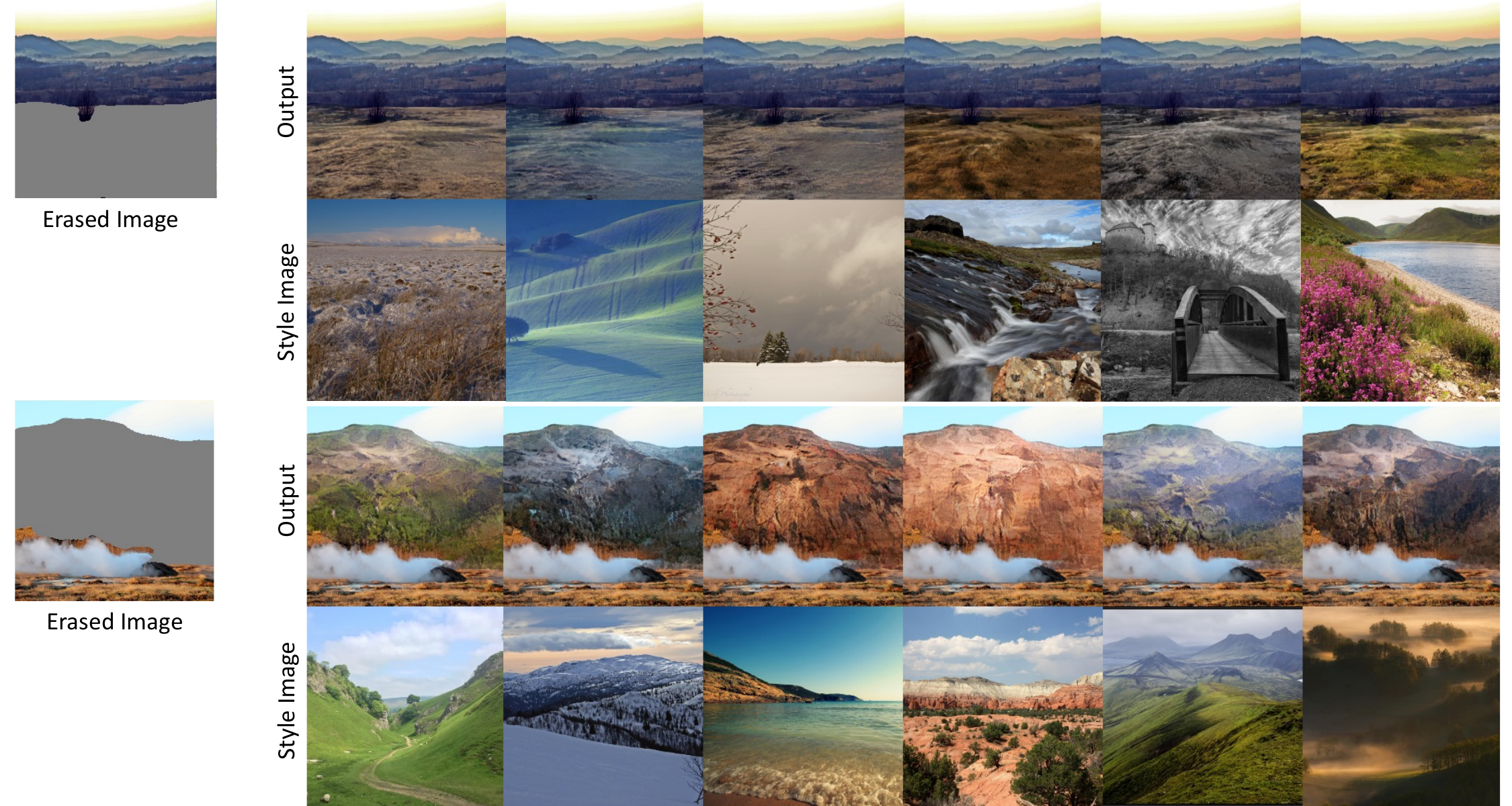}
\caption{Diverse editing examples of our method. The erased image is inpainted with different style codes that are extracted from corresponding semantic classes of the style images.}
\label{fig:diverse_scene}
\end{figure*}

Qualitative results of replacing all objects that belong to the same semantic category or adding a single instance examples are shown in Fig. \ref{fig:comparisons_add_ade} on ADE20k images.
These visuals are obtained for examples replacing all car classes in the scene or adding a single instance of a car to the scene.
In Fig. \ref{fig:comparisons_add_ade}, our method is the only one that can incorporate style information for each instance and it affects especially the examples where we add multiple instances to a scene.
For example, the third column from Fig. \ref{fig:comparisons_add_ade} shows adding the curtain class with two instances as shown with different color codes. 
Other methods struggle to generate realistic two-piece curtains and the texture is muddled. 
On the other hand, our results are sharp, and two-piece curtains have clear boundaries and styles.
Similar behavior can be observed in other examples as well.
Note that we do not compare with SIEDOB on ADE20k images because the model has not been released yet. 
However, SIEDOB considers the bed, chest, lamp, chair, and table as foreground objects and handles curtains as background. 
Therefore SIEDOB also would not output diverse styles for these examples.

In Fig. \ref{fig:comparisons_add_city}, we compare all methods on object addition and semantic replacement task on Cityscapes images.
We observe that when the object addition is provided as a rectangle box as in the first example, SIEDOB outputs lower quality results. This also shows in the FID scores of SIEDOB for adding objects in Table \ref{table:results_all}.
Even though SIEDOB is slightly better than SPMPGAN in the majority of the settings, in adding objects when we provide those objects with the smallest rectangle covering the instance, it performs worse than SPMPGAN.
We also observe that SIEDOB can generate very good car instances when they are not occluded but not as good when the cars occlude each other as in the example second row of side-parking cars. 
Our method achieves better results for those examples than presented competing works.

Lastly, in our user study, users select our method as opposed to SPMPGAN $65\%$ of the time ($50\%$ is a tie) because ours achieves better color consistency and better instance generations. These results are consistent with the FID metrics from Table \ref{table:results_all}.

\begin{figure*}[]
\includegraphics[width=\textwidth]{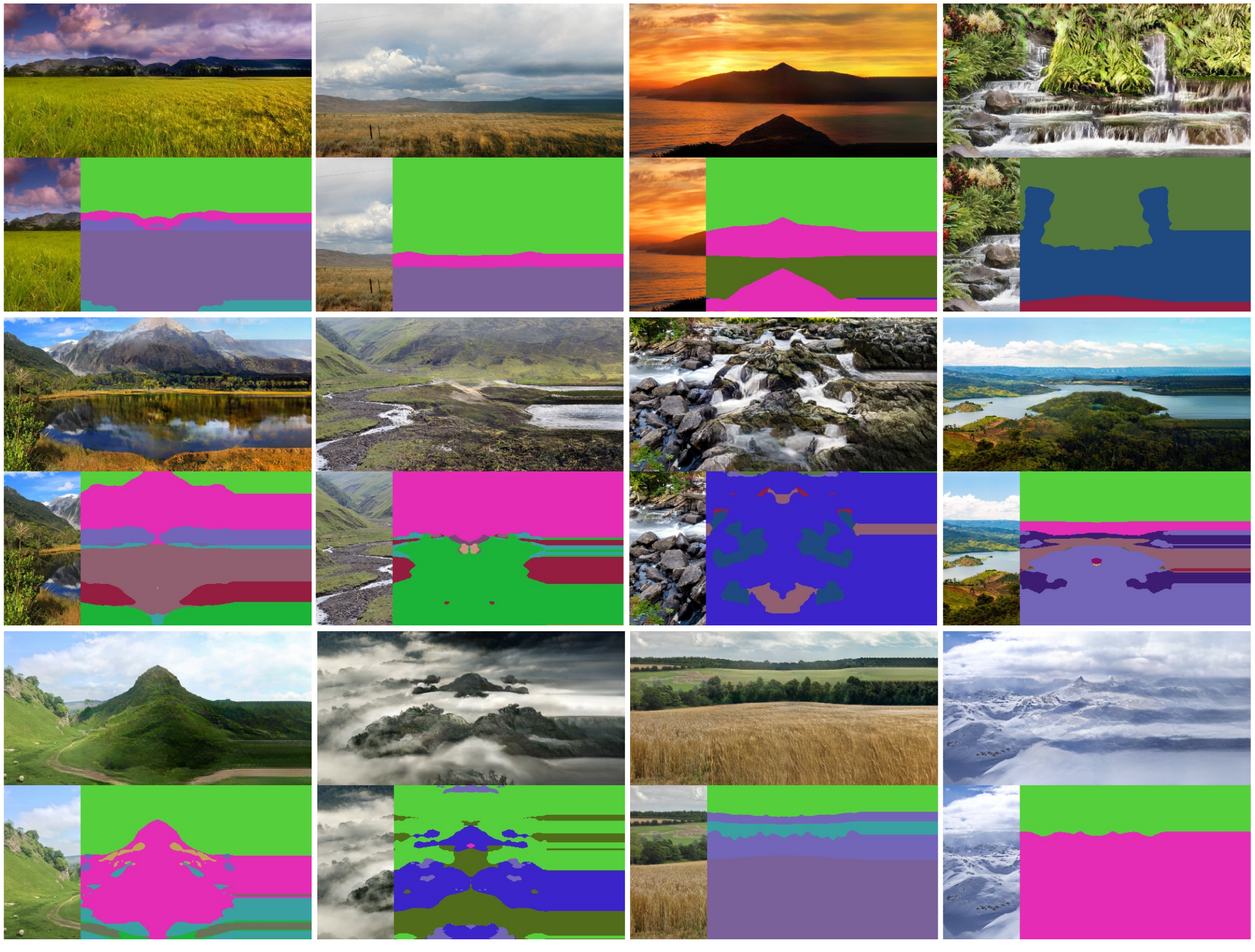}
\caption{Panorama results of our method. With the given semantic map, we can extend the images.}
\label{fig:panorama}
\end{figure*}

\subsection{Ablation Study}
We provide an ablation study in Table \ref{table:ablation} that is conducted on ADE20k-Landscape since it is a smaller dataset and the training is faster.
We track the FID scores which shows the quality of the output images.

We first report the results that do not have the style encoding. Removing the style encoding mechanism altogether corresponds to the SPMPGAN model and it achieves worse results and it is deterministic. 
Next, we validate the effectiveness of the style encoding that considers the valid ratio as explained in Section \ref{sec:style_encoding}.
Our results show that using the mechanism that takes the valid ratio and outputs shift and scale parameters for the style improves the results.
Next, we experiment if the multi-scale generator is useful in our framework. For these experiments, we only train the Gen3 architecture that takes the input as the original resolution.
Its result is worse than the final model.
We also run an ablation study for different dimensions of styles we extract via style encoding. 
This dimension is set via the last channel size of the style encoder network.
The results are close but we find that style dimension $128$ achieves the best score. 
The smaller dimension may not be enough for the style encoding and the larger dimension may carry too much information resulting in overfitting.

\subsection{Applications}
The proposed work achieves multiple image applications under one framework. 
We can remove objects from the scene and add new objects. The most important difference between our work and previous ones is that we can add new objects with diverse styles. 
Examples of these results are shown in Fig. \ref{fig:diverse_objects}.
Similarly, we can change the style of a semantic category as shown in Fig. \ref{fig:diverse_scene}.
We share the output results and the images that we take the style from for the corresponding semantic category.
As can be seen from Fig. \ref{fig:diverse_scene}, the output semantic styles resemble the semantic regions from the style image.

Finally, our method is able to generate panoramas by recursively outpainting an image as shown in Fig. \ref{fig:panorama}. 
For these visual results from Fig. \ref{fig:panorama}, we collect images from Flickr to share more interesting outputs. We have a total of 7389 images, which we divided into training and validation sets, with 6651 images in the training set and 738 images in the validation set.
We use UperNet to segment the images and train a model on this dataset. 
It provides more variety in the landscapes compared to the ADE20k-Landscape dataset which we used to compare with previous works.
 Fig. \ref{fig:panorama} shows images from the validation set. 
 To generate realistic segmentation maps, we first extend the segmentation map by mirroring and then by repeating the border pixels.
 The outputs look good quality and realistic with many details.
 We also provide a demo of the model that is trained on the Flickr dataset. The demo can be found on the web page provided in the Abstract.

\section{Conclusion}

We present a novel framework for semantic image editing that achieves object removal, diverse object addition and semantic inpainting, and semantic map-conditioned panorama generation under one roof.
Our framework includes a method that encodes visible and partially visible objects via a mask-aware style encoding module.
We extensively compare with previous conditional image generation and semantic image editing algorithms. 
Our extensive experiments show that our method significantly improves over the state-of-the-art and not only achieves better performance than previous works but also provides diverse results.

\subsection*{Acknowledgement}

This work has been funded by The Scientific and Technological Research Council of Turkey (TUBITAK), 3501 Research Project under Grant
No 121E097. 

{
\bibliographystyle{ieee}

}



\end{document}